# *Testing and Debugging Techniques for Answer Set Solver Development*

ROBERT BRUMMAYER

*Institute for Formal Models and Verification, Johannes Kepler University Linz, Austria*

MATTI JÄRVISALO

*Department of Computer Science, University of Helsinki, Finland*

## Abstract

This paper develops automated testing and debugging techniques for answer set solver development. We describe a flexible grammar-based black-box ASP fuzz testing tool which is able to reveal various defects such as unsound and incomplete behavior, i.e. invalid answer sets and inability to find existing solutions, in state-of-the-art answer set solver implementations. Moreover, we develop delta debugging techniques for shrinking failure-inducing inputs on which solvers exhibit defective behavior. In particular, we develop a delta debugging algorithm in the context of answer set solving, and evaluate two different elimination strategies for the algorithm.

## 1 Introduction

Answer set programming (ASP) (Gelfond and Lifschitz 1988; Niemelä 1999) is a rule-based declarative programming paradigm that has proven to be an effective approach to knowledge representation and reasoning in various hard combinatorial problem domains. This success has been enabled by the development of efficient answer set solvers (Simons et al. 2002; Ward and Schlipf 2004; Lin and Zhao 2004; Janhunen and Niemelä 2004; Liu and Truszczynski 2005; Anger et al. 2005; Leone et al. 2006; Giunchiglia et al. 2006; Janhunen 2006; Gebser et al. 2007; Drescher et al. 2008; Janhunen et al. 2009; Brain and De Vos 2009).

Implementing robust, sound and complete answer set solvers is a demanding task. For achieving high solver performance, one needs to implement error-prone and complex inference rules, specialized data structures, and other complex optimizations. On the other hand, robustness and correctness are two essential criteria for answer set solvers. The users of answer set solvers heavily depend on correct results and, in particular, correct answer sets. The lack of systematized testing tools for answer set solver development may leave intricate implementation bugs unnoticed. Indeed, in practice, small sets of problem instances that are typically used during regression and unit testing are not enough for testing correctness during solver development. Moreover, while the availability of standard benchmark instances is of high importance for benchmarking solver implementations, testing during solver development should not solely rely on these instances. In support of these claims, by examining the detailed results of the first and second ASP programming competitions (Gebser et al. 2007; Denecker et al. 2009) one notices that, on the sets of (typical) benchmarks used in these competition, only very few solvers on very few benchmarks were judged as providing incorrect results. In other words, almost no defective behavior seems to have been detected. In contrast, we will show that by using the testing and debugging techniques developed in this work, various kinds of incorrect and erroneous behavior can be automatically detected and debugged for various state-of-the-art answer set solvers; furthermore, this is achieved without the need for the user to construct hand-crafted benchmarks. The testing techniques developed here provide complementary means for developing highly correct solvers, which can be applied in addition to domain-specific benchmarks. Additionally, our delta debugging techniques can be naturally applied also when defective solver behavior is detected on domain-specific benchmarks.

In more detail, this paper develops domain-specific *grammar-based black-box fuzz testing* and *delta debugging* techniques that enable more systematic testing and debugging solutions for answer set solver development. Fuzz testing, also called *fuzzing* (Sutton et al. 2007; Takanen et al. 2008), has its origin in software security and quality assurance. The main idea of fuzzing is to test software against random inputs in order to find *failure-inducing inputs* that trigger defective behavior. In order to find as many defects as possible, a "good" *fuzzer* (the input generator) should generate a wide variety of different inputs. In the *grammar-based* approach, the generated input is guaranteed to be *syntactically* valid, i.e. the input respects the expected input format. In *black-box fuzzing*, testing is performed against the software interface without access to the implementation details of the software. We develop a black-box grammar-based fuzz testing tool that is able to generate random ASP instances from various different classes.

In many cases, randomly generated failure-inducing inputs may contain large parts that are irrelevant for triggering defective behavior, and can hence be too large to enable efficient debugging. In the context of ASP, a failure-inducing answer set program can simply have too many rules and atoms for a developer to manually pinpoint the subset of program rules that triggers the defective behavior. In order to isolate the failure-inducing parts of such failure-inducing inputs, an automatic technique called *delta debugging* (Zeller and Hildebrandt 2002; Zeller 2005; Claessen and Hughes 2000; Misherghi and Su 2006) has been proposed. Minimization of the size of failure-inducing input is intractable in general, and hence delta debuggers are based on greedy heuristics. We develop an efficient and novel delta debugger which is very effective in producing small failure-inducing inputs (answer set programs), hence relieving the solver developer from the troublesome task of locating the typically small failure-inducing parts of a large failure-inducing answer set program.

*Main contributions* We develop a *grammar-based black-box fuzz testing tool for answer set solvers* that covers various different classes of grounded answer set programs. Our experimental analysis shows that our fuzzing approach is very effective in revealing various sources of incorrectness, including both unsound and incomplete behavior, in a wide range of state-of-the-art answer set solvers. Additionally, of independent interest is that since our fuzzer is highly configurable, it can also be used as a flexible ASP benchmark generator. Moreover, we develop a novel delta debugging algorithm for answer set solver development. Our algorithm uses the hierarchical (Misherghi and Su 2006) structure of answer set programs to shrink failure-inducing instances effectively. Furthermore, we evaluate two different elimination strategies we have implemented for our delta debugging algorithm: a simple and easy-to-implement "one-by-one" elimination strategy, and another one based on the more intricate DDMin algorithm originally proposed in different context (Zeller and Hildebrandt 2002).

All tools developed in this work are publicly available and can be downloaded at `http://fmv.jku.at/fuzzddtools/`. Since the tools treat answer set solvers to be tested as black-box entities, no modifications to the actual solvers are needed in order to apply these tools in the development process of any answer set solver that accepts input in the standard *lparse* format.

The rest of this paper is organized as follows. First we review necessary concepts related to answer set programs (Section 2). We then present our fuzzing approach for testing answer set solvers (Section 3), with experimental results of the effectiveness of the fuzzer (Section 4). Finally, we develop delta debugging strategies in the context of answer set solving (Section 5) and present an experimental evaluation of our approach (Section 6), followed by pointers to related work (Section 7).

## 2 Answer Set Programs

This section reviews the stable model semantics and the classes of answer set programs considered in this work.

*Normal Logic Programs and Answer Sets* A normal logic program (NLP) consists of a finite set of *normal* rules of the form

$$r \ : \ h \leftarrow a_1, \ldots, a_n, \sim b_1, \ldots, \sim b_m,$$

where each $a_i$ and $b_j$ is a propositional (or *normal*) atom, and $h$ is either a propositional atom or the symbol $\bot$ that stands for falsity. A rule $r$ consists of a *head*, $\mathsf{head}(r) = h$, and a *body*, $\mathsf{body}(r) = \{a_1, \ldots, a_n, \sim b_1, \ldots, \sim b_m\}$. A rule $r$ is a *fact* if $\mathsf{body}(r) = \emptyset$, and an *integrity constraint* if $head(r) = \bot$. The symbol "$\sim$" denotes *default negation*. A *default literal* is an atom $a$ or its default negation $\sim a$.

For a rule $r$, let $\mathsf{body}(r)^+ = \{a_1, \ldots, a_n\}$ and $\mathsf{body}(r)^- = \{b_1, \ldots, b_m\}$ denote the sets of positive and negative (default negated) atoms in $\mathsf{body}(r)$, respectively.

In ASP, we are interested in *stable models* (Gelfond and Lifschitz 1988) (or *answer sets*) of a program $\Pi$. An *interpretation* $M \subseteq \mathsf{atom}(\Pi)$ defines which atoms of $\Pi$ are true ($a \in M$) and which are false ($a \notin M$). An interpretation $M \subseteq \mathsf{atom}(\Pi)$ satisfies a normal rule $r$ if and only if $\mathsf{body}(r)^+ \subseteq M$ and $\mathsf{body}(r)^- \cap M = \emptyset$ imply $\mathsf{head}(r) \in M$, and hence $M$ is a *(classical) model* of $\Pi$ if $M$ satisfies all rules in $\Pi$.

A model $M$ of a program $\Pi$ is an answer set of $\Pi$ if and only if there is no model $M' \subset M$ of $\Pi^M$, where $\Pi^M = \{\mathsf{head}(r) \leftarrow \mathsf{body}(r)^+ \mid r \in \Pi \text{ and } \mathsf{body}(r)^- \cap M = \emptyset\}$ is called the *Gelfond-Lifschitz reduct* of $\Pi$ with respect to $M$. The problem of deciding whether a NLP has an answer set is NP-complete.

*Weight Constraint Programs* In order to enable more convenient modeling in ASP, extensions of normal programs have been proposed. Examples of such extensions are what we refer here to as *weight constraint programs* (WCPs). A *weight atom* is of the form

$$l[a_1 = w_{a_1}, \ldots, a_n = w_{a_n}, \sim b_1 = w_{b_1}, \ldots, \sim b_m = w_{b_m}]u,$$

where each $l_i \in \{a_i\}_{1 \leq i \leq n} \cup \{\sim b_j\}_{1 \leq j \leq m}$ is a default literal, each $w_i$ an integer (the *weight* of $l_i$), and $u, l$ are integers with $l \leq u$ (the *lower* and *upper bound*, respectively). If one of the bounds $u$ and $l$ is omitted, this bound is implicitly $\infty$ (for $u$) or $-\infty$ (for $l$). A *cardinality atom* is the special case of a weight atom in which each literal has weight one. The variant of cardinality atoms in which *both* of $u, l$ are omitted is called a *choice atom*, that is an expression of the form $\{a_1, \ldots, a_n\}$, where each $a_i$ is a normal atom.

A weight constraint rule is of the form $r : C_0 \leftarrow C_1, \ldots, C_n$, where the head $C_0$ is a normal, weight, cardinality, choice atom, or $\perp$ and each $C_i$, $i > 0$ in the body (which can also be empty) is a normal, weight, or cardinality literal. We use the term *weight constraint programs* for the set of programs that consist of weight constraint rules. Hence NLPs are special cases of weight constraint programs.

Given an interpretation $M$, the stable model semantics extends to weight constraint programs by defining that a weight atom is satisfied by $M$ if and only if $l \leq \sum_{a_i \in M} w_i + \sum_{b_j \notin M} w_j \leq u$. A choice atom is always satisfied by $M$. The problem of deciding whether a given weight constraint program has an answer set remains in NP.

*Disjunctive Logic Programs* Another extension of normal rules are *disjunctive rules*, in which the head can, instead of a normal atom, be a disjunction $\bigvee_{i=1}^{n} a_i$ of normal atoms. Disjunctive logic programs (DLPs) can contain normal and disjunctive rules. The stable model semantics extends to the disjunctive case naturally by defining that a disjunctive rule $r$ with $\mathsf{head}(r) = \bigvee_{i=1}^{n} a_i$ is satisfied by an interpretation $M$ if and only if $\mathsf{body}(r)^+ \subseteq M$ and $\mathsf{body}(r)^- \cap M = \emptyset$ imply $a_i \in M$ for some $i$. The problem of deciding whether a DLP has an answer set is $\Sigma_2^p$-complete and thus presumably harder than the case of NLPs.

## 3 Fuzz Testing Answer Set Solvers

In this section we develop a native grammar-based black-box fuzzing approach for testing answer set solvers.

### 3.1 Grammar-Based ASP Fuzzing

In order to apply grammar-based fuzz testing to answer set solvers, methods for generating wide varieties of different answer set programs need to be developed. There are only a few studies that consider the problem of generating random logic programs in the context of ASP (Zhao and Lin 2003; Namasivayam and Truszczyński 2009). These studies consider rather restricted subclasses of NLPs and focus on theoretical aspects such as the study of phase transition behavior. In contrast, our aim here is to generate a wide variety of different random answer set programs in order to *test* answer set solvers.

### 3.1.1 Ineffectiveness of CNF-Based ASP Fuzzing

A simple approach to generating random answer set programs consists of first using generators for random conjunctive normal form (CNF) instances of the Boolean satisfiability (SAT) problem and translating the generated CNFs into answer set programs afterwards. However, this approach appears to be ineffective, as revealed by the following evaluation.

We obtained CNF instances by generating random propositional formulas as Boolean circuits and translating them to CNF via a standard encoding (Tseitin 1983). The NLPs were obtained from the CNF instances using the following standard translation: given a CNF $F$, introduce (i) for each Boolean variable $x$ in $F$ the rules $x \leftarrow \sim\hat{x}$ and $\hat{x} \leftarrow \sim x$ (forcing classical interpretations); and (ii) for each clause $c$ in $F$, the rule $\bot \leftarrow \sim c$ and for each Boolean variable in $c$ the rule $c \leftarrow x$ ($c \leftarrow \sim x$, resp.) if $x$ occurs positively (negatively, resp.) in $c$ (stating the clause should be satisfied). Notice that this translation always results in *tight* NLPs, a subclass of NLPs.

Notably, using a 1-hour time limit and the same hardware settings as in our latter experiments, we tested all of the answer set solvers that are shown in Table 1 on 8850 CNF instances, but did not find any defects. We conjecture that the CNF-based fuzz testing approach for ASP is unsuccessful as it lacks domain knowledge and considers only tight NLPs. This gives motivation to develop *domain-aware* fuzz testing approaches for answer set solver development, which take the specific features of different ASP classes into account.

### 3.2 FuzzASP: A Native ASP Fuzzer

In order to generate a wide variety of different answer set programs, we developed FuzzASP, which is a native fuzzer for ground answer set programs generated in the syntax of lparse. In addition to normal logic programs, it supports combinations of disjunctive and extended rules with choice, cardinality and weight atoms, and classical negation. FuzzASP is able to generate varying types of random program instances from large classes of programs in order to provide high variety for different combinations of rule constructs.

FuzzASP generates programs as follows. Let $A$ be a set of $n$ normal atoms.

1. A set of $f$ facts (normal rules with empty bodies) is generated by picking each head uniformly at random (u.a.r.) from $A$.
2. Normal rules with non-empty bodies and varying lengths are generated until each atom in $A$ occurs in at least $r_b$ and $r_h$ bodies and heads, respectively. Each normal rule is generated by picking the head and each body atom u.a.r.

from $A$. Moreover, each body atom is default negated with probability $p_{dn}$. The body length of each normal rule is chosen u.a.r. from a predefined range.
3. A set of $i$ integrity constraints is generated, picking each body atom u.a.r. from $A$ and default negating each atom with probability $p_{dn}$.

When generating a WCP:

4. A set of $W$ weight constraint rules is generated. The head of each rule is randomly chosen to be a normal atom from $A$, or a weight, cardinality, or choice atom. A weight atom $l[a_1 = w_{a_1}, \ldots, a_n = w_{a_n}, \sim b_1 = w_{b_1}, \ldots, \sim b_m = w_{b_m}]u$ is generated by picking atoms u.a.r. from $A$, and negating each atom with probability $p_{dn}$. The weights for the literals and the bounds $l$ and $u$, where $l \leq u$, are chosen randomly. The number of normal atoms to appear in each weight atom is chosen u.a.r. from a predefined range. Additionally, one of the bounds $u$ and $l$, and weights of individual literals are left out with certain probabilities. Cardinality and choice atoms are generated analogously. Each body literal is similarly chosen to be a normal, weight or cardinality atom.

When generating a DLP:

4. A set of $d$ disjunctive rules are generated. A disjunctive head is generated by picking $d_a$ atoms u.a.r. from $A$ (similarly for the normal atoms in the bodies, default negating with probability $p_{dn}$). The head length of each disjunctive rule is chosen u.a.r. from a predefined range.

The fuzzer has been designed to be highly configurable. Nearly every detail can be configured. However, this is optional as the fuzzer already comes with reasonable default values. Due to page limitations, for details on the actual default values provided in the implementation of FuzzASP, please refer to the help provided in the actual FuzzASP implementation.

We have configured the default values through experimentation so that the generated logic programs are not trivial but also not too hard to solve. One key success factor of fuzz testing is high test throughput, which means that generating hard instances solely is counterproductive. On the other hand, trivial instances are unlikely to be critical failure-inducing inputs, as they can often be solved in early phases of the answer set solver, e.g. in a pre-processing phase. Therefore, in order to generate various different programs of varying difficulty, the fuzzer randomizes its parameters. For each parameter a minimum and a maximum value is considered. The fuzzer respectively picks one value within the particular range.

### 3.3 Solver Defect Categories

We divide defects of answer set solvers into three categories:

**Errors** contains instances on which the solver terminates in an unexpected way without providing a result, e.g. segmentation faults and assertion failures;
**Invalid models** contains instances on which the solver reports a solution that is not an answer set (either not minimal or not even a classical model) of the instance; and **Incorrect** where, if the solution provided by at least one solver is a correct answer set, all solvers that report that no answer sets exist are treated as incorrect.

Notice that these categories are disjoint in the sense that, for a given instance, each solver can only exhibit a defect that belongs to exactly one of the categories. Notice also that in this paper we concentrate on the problem of answer set existence. Hence, especially considering the defect categories *invalid models* and *incorrect*, for each solver a single answer set for each instance is checked for correctness. However, the testing and debugging techniques developed in this work can also easily be adapted for the problem of answer set enumeration, that is, for checking the validity of all answer sets reported by a solver.

Considering the defect category *invalid models*, for a given instance $\Pi$, we employ the following method for checking if solutions reported by solvers are valid answer sets of $\Pi$. For given a model candidate $M$ reported by a solver, we construct the set $I_M = \bigcup_{a \in M}\{\bot \leftarrow \sim a\} \cup \bigcup_{a \notin M}\{\bot \leftarrow a\}$ of integrity constraints. Then $M$ is a valid answer set of $\Pi$ if and only if $\Pi \cup I_M$ has an answer set. Notice that it is trivial to check whether $\Pi \cup I_M$ has an answer set. Using this method, also *incorrect* solver behavior is captured in the case a solver claims that there are no answer sets for a given instance $\Pi$, if some other solver reports an answer set that is determined as a valid one using the checking method.

Furthermore, we also crosscheck all occurrences in the categories *invalid models* and *incorrect* using voting. In more detail, assume that, based on the above-described checking method, a specific solver $S$ reports an invalid model (or claims incorrectly that no answer sets exist) for a given instance $\Pi$. Then, by running a set of solvers on $\Pi$, we crosscheck invalid models and incorrectness by checking that a majority of the solvers report that there are no answer sets for $\Pi$ (or report an answer set), that is, a majority of the solvers vote against the output of the solver $S$ on $\Pi$. In the ideal case, all other solvers vote against the output of $S$. In our experiments in this paper, all crosschecks turned out to be ideal in this sense.

## 4 Fuzzing Experiments

We performed fuzz testing experiments using FuzzASP for the following classes of logic programs: NLP (normal programs), WCP (weight constraint programs), and DLP (disjunctive programs). We ran our experiments under Ubuntu Linux on an Intel Core 2 Quad 2.66 GHz machine with 8 GB of RAM. Our fuzzing framework used all the four cores for parallel testing. Using default settings, we tested a wide selection[1] of answer set solvers that participated in the first (Gebser et al. 2007) or second (Denecker et al. 2009) ASP Competition in 2007/2009. The grounder lparse (version 1.1.1) was used as a front-end for the solvers. The only exception was the solver DLV, which does not require an external front-end. For each class, we restricted the total fuzz testing time to one hour.

---

[1] Solvers: Clasp 1.2.1, ClaspD 1.1, Cmodels 3.79, DLV precompiled build BEN/Oct 11 2007, GnT2 precompiled v. 2.1 using Smodels 2.33 as backend, lp2diff precompiled 1.19 with lp2normal 1.7 using Z3 2.0 SMT solver (de Moura and Bjørner 2008) as backend, lp2sat precompiled 1.11 with lp2atomic 1.12 using Picosat 913 SAT solver (Biere 2008) as backend, noMoRe++ 1.5., PBmodels 0.2 using Minisat+ 1.0 pseudo-boolean solver as backend, Smodels 2.33, Smodels-ie standalone 1.0.0, Smodels_cc 1.08, SUP 0.4 using Minisat 1.12b SAT solver (Eén and Sörensson 2004).

We want to emphasize that our goal is not to present results for *all* available solvers, but rather to demonstrate the wide applicability of our testing and debugging techniques on different types of answer set solvers.

The experimental results for NLP, WCP, and DLP are presented in Tables 1, 2, and 3, respectively. In short, our fuzz testing approach is very effective in finding solver defects in state-of-the-art answer set solvers, due to the impressive number of critical defects found in the experiments. Next, we will discuss the results for each of the considered program classes (NLP, WCP, and DLP) in more detail. Full results including failure-inducing inputs can be found in the archive http://fmv.jku.at/brummayer/fuzz-dd-asp.tar.7z.

### 4.1 Defects Found on NLP

For the class NLP, the 1-hour time limit resulted in testing the solvers listed in Table 1 on 10190 instances generated by FuzzASP. For generating NLPs, we used the default options of FuzzASP (with weight and cardinality literals disabled). As shown in Table 1, the effectiveness of detecting solver defects in NLP is rather modest. Mostly, errors such as crashes were detected, most notably in high numbers for lp2sat, Smodels-ie, and SUP. A few invalid models were reported by Cmodels and lp2diff. Moreover, we found two instances on which Clasp incorrectly reports that there are no answer sets.

Table 1. NLP fuzzing results on 10190 instances.

| solver          | errors | invalid models | incorrect |
|-----------------|--------|----------------|-----------|
| Clasp           | 0      | 0              | 2         |
| Cmodels         | 1      | 3              | 0         |
| DLV             | 0      | 0              | 0         |
| lp2sat(picosat) | 3221   | 0              | 0         |
| lp2diff(z3)     | 0      | 6              | 0         |
| noMoRe++        | 0      | 0              | 0         |
| Pbmodels        | 0      | 0              | 0         |
| Smodels         | 0      | 0              | 0         |
| Smodels-ie      | 1635   | 0              | 0         |
| Smodels_cc      | 5      | 0              | 0         |
| SUP             | 1690   | 0              | 0         |

This is already in contrast to the ineffective CNF based ASP fuzzing experiment, confirming our conjecture that domain-unaware fuzz testing approaches are in general *not* effective in finding solver defects, and therefore domain-aware fuzzing techniques have to be developed individually.

In order to verify the validity of a model $M$ reported by a solver for a test instance $\Pi$, we used Smodels as *trusted solver* for checking whether the program $\Pi \cup I_M$ has an answer set (the original test instance enhanced with the model candidate as integrity constraints $I_M$). This check is trivial as it requires only deterministic propagation and no search. Smodels was chosen for NLP and additionally for WCP, since it exhibited neither errors nor incorrect results.

We want to further stress that although we do not have the precise running time data for individual solvers available, the total time for this experiment was one hour wall clock time using four processor cores, and hence the testing time used on each of the 11 solvers was around 20 minutes on average, and in this time over 10000 test cases were tried, which totals in over 110000 solver calls. Furthermore, we would like to point out that the 1-hour testing time limit used in the fuzzing experiments was only enforced for obtaining a representative snapshot to the effectiveness of the testing technique. In practice, the testing loop works by generating one test case at a time, and running all solvers on this test case. For testing a specific solver, one can stop the testing loop as soon as a single failure (error, incorrect result, or the like) is detected for the specific solver.

### *4.2 Defects Found on WCP*

In particular, the effectiveness of FuzzASP is very impressive on WCP and DLP. The fuzz testing results on 19840 inputs for WCP are shown in Table 2. The input logic programs were generated using the FuzzASP option of introducing additional choice, cardinality and weight rules, each up to 5% of generated normal rules. Here, we tested those solvers that accept the class WCP (supported by lparse) as input.

Table 2. WCP fuzzing results on 19840 instances.

| solver | errors | invalid models | incorrect |
|---|---|---|---|
| Clasp | 0 | 2 | 6 |
| Cmodels | 2004 | 7 | 78 |
| lp2diff(z3) | 0 | 6 | 2 |
| Smodels | 0 | 0 | 0 |
| Smodels-ie | 1651 | 16 | 11 |
| SUP | 2224 | 1 | 71 |

The results on NLP and WCP suggest that many defects are due to the techniques implemented for inference on weight constraint rules. Based on the results, Smodels appears to be the most stable solver for this class of programs, being the only solver for which no defects were found. As an example of the difficulty of maintaining correctness while optimizing solver performance, we observed a high number of defects in each category for Smodels-ie, which is a re-implementation of Smodels with improved data structures aimed at better memory locality.

For answer set solvers that apply different back-end solvers such as SAT and SMT solvers, the back-end solver may be to blame for incorrectness. Taking lp2diff as an example, we did a cross-check in order to pinpoint the source of incorrectness on both NLP and WCP. The same incorrect behavior occurred when two other SMT solvers, CVC3 (Barrett and Tinelli 2007) and Yices (Dutertre and de Moura 2006) were used as back-end solvers. Hence, the source of incorrectness is highly likely to be in lp2diff itself.

Again, notice that the total time for this experiment was one hour wall clock time (including all setup and test instance generation times) using four processor cores, and hence the testing time used on each of the six solvers was around 40 minutes on average, and in this time close to 20000 test cases were tried.

### 4.3 Defects Found on DLP

The fuzz testing results on DLP are shown in Table 2. Here we used the FuzzASP option of introducing disjunctive rules up to 5% of generated normal rules.

While no defects were found for DLV and GnT2, a vast number of defects were found for ClaspD and Cmodels. Due to its robustness, DLV was used as trusted solver for checking validity of reported models in DLP. Based on these results, we conclude that many defects are due to the techniques particularly implemented for handling disjunctive rules.

Table 3. DLP fuzzing results on 33050 instances.

| solver  | errors | invalid models | incorrect |
|---------|--------|----------------|-----------|
| ClaspD  | 9      | 255            | 30        |
| Cmodels | 806    | 3366           | 28        |
| DLV     | 0      | 0              | 0         |
| GnT2    | 0      | 0              | 0         |

## 5 Delta Debugging for Answer Set Solvers

Now, we focus on developing delta debugging algorithms for answer set solvers. The overall goal of delta debugging (Zeller and Hildebrandt 2002; Zeller 2005; Claessen and Hughes 2000; Misherghi and Su 2006) is to minimize the size of failure-inducing inputs while maintaining the same observable behavior. In this way, large irrelevant parts of the inputs are pruned away, resulting in small program instances that consist of isolated failure-inducing parts.

### 5.1 The Delta Debugging Algorithm

As an overview, our delta debugging algorithm DeltaASP works as follows. The delta debugger runs the solver on the original failure-inducing input in order to observe the defective behavior, e.g. the solver crashes with a segmentation fault, or outputs a solution which is not a valid answer set. Then, the delta debugger iteratively tries to eliminate parts of the current input. After each elimination, the delta debugger runs the solver on the current (reduced) input. If the solver shows the same observable behavior, the delta debugger continues with this reduced input. Otherwise, the delta debugger undoes the last elimination, and continues with other eliminations. Finally, after a given time limit or after reaching a fix-point, the delta debugger terminates and outputs a smaller program that is guaranteed to trigger the same observable behavior as the original program instance.[2]

Notice that the goal of delta debugging is to obtain a *small* failure-inducing input within a reasonable time limit, e.g. a few seconds or minutes. In other words, in practice, engineers are hardly interested in minimal failure-inducing inputs if they have to wait a long time. Therefore, delta debuggers typically apply (greedy) elimination heuristics for reducing failure-inducing inputs within a small time limit.

Given a failure-inducing answer set program $\Pi$ as input, the eliminations attempted by the DeltaASP delta debugging algorithm can be divided into the following phases:

1. Remove rules from $\Pi$ until fix-point (heuristically).
2. For each rule $r \in \Pi$: if $r$ is neither a fact nor a constraint, then try to replace head($r$) with $\bot$ and resp. body with $\emptyset$.
3. If at least one rule could be reduced in phase 2, goto 1.
4. For each rule $r \in \Pi$: try to remove individual literals from head($r$) resp. body($r$) while |head($r$)| > 1 resp. |body($r$)| > 1.
5. For each rule $r \in \Pi$: try to remove individual elements of each weight, cardinality, and choice literal in $r$ while elements are left.
6. For each rule $r \in \Pi$: try to remove the negation from individual negative literals in $r$.
7. If at least one rule could be reduced in 4–6, goto 1. Otherwise, output the current program and terminate.

DeltaASP can be seen as a variation of hierarchical delta debugging (Misherghi and Su 2006), since our method proceeds from the top-most elements of the hierarchy (rules) to lower-level elements: first rules, then individual heads and bodies of rules, then individual literals, and, at last, negations. As a greedy heuristic, our *primary objective* is to minimize the number of rules as soon as possible. This may drastically prune large irrelevant parts of the inputs up front. As removing and reducing individual rules may enable removing rules that could not be removed before, we perform rule removal until fix-point in phase 1, and perform a "restart" in phase 3 if at least one rule could be reduced in phase 2. We perform this restart in order to minimize the number of rules, heads and bodies up front before we try more fine-grained reductions in phases 4–6. Typically, as soon as we reach phase 4, the input has already been reduced significantly.

Our *secondary objective* is to reduce individual rules after no more rules can be removed. These reductions are performed in phases 2 and 4–6. Again, reducing individual rules may enable the removal and reduction of rules that could not be removed resp. reduced before. As restarting ("goto 1") after each individual phase of 4–6 can be too costly, we postpone restarts until phase 7.

---

[2] For the defect categories *invalid models* and *incorrect*, *same observable behavior* is checked against the result reported by the trusted solver on the same instance. Another possibility would be to employ majority voting by running multiple solvers on the same instance.

*5.2 Removal strategies*

In phases 1, 4, and 5 we consider a set from which we want to eliminate as many elements as possible, e.g. the set of rules in phase 1. Next, we discuss and evaluate the differences between a simple *one-by-one* approach (OBO) and a more intricate strategy based on the DDMin algorithm (Zeller and Hildebrandt 2002).

*DDMin*  The original DDMin algorithm (Zeller and Hildebrandt 2002) attempts to divide the current set into $k$ subsets, where $k$ (the *granularity*) is initialized to 2. If at least one of the subsets is enough to reproduce the same observable behavior, the current set is reduced to this subset, granularity is reset to 2, and the algorithm continues. Otherwise, it tries the complement sets of each of the subsets. If using the set complements does not succeed either, the granularity $k$ is doubled, i.e. in the next iteration the current set is divided into smaller subsets. In the last iteration, the granularity is equal to the size of the current set, which means that each element is in its own subset. Notice that in order to avoid recomputations on already considered subsets, intermediate results need to be cached.

If the failure-inducing input part is rather local and does not depend much on other input parts, the DDMin algorithm tends to simulate a binary search strategy during the first iterations, since the current set can often be reduced to one of its considered two subsets. However, if the failure-inducing part strongly depends on other parts of the input, trying $k$ subsets up front will only seldom lead to success. Then, DDMin has to consider the set complements and to iteratively increase the granularity, which may be rather ineffective.

*OBO*  As an alternative to DDMin, we also consider a simple strategy based on a *one-by-one* (OBO) principle. We iterate over all elements in the set and try to remove them one by one. After each iteration, we repeat the process if at least one element could be removed. In principle, we could immediately restart the algorithm as soon as we have been able to remove one element. However, this may be too costly and therefore the restart is postponed until the end of the iteration. The benefit of the OBO strategy is that, in contrast to the DDMin strategy, it is easy to implement and does not need any caching techniques.

## 6 Delta Debugging Experiments

For the delta debugging experiments, we used the same hardware (this time using a single processor core) and settings as for the fuzzing experiments. However, the delta debugging experiments were not run simultaneously. For the experiments, we used the failure-inducing inputs found in the fuzzing experiments reported in Section 4. Depending on the classes of errors and error messages, we semi-automatically divided the failure-inducing inputs into different bug classes.

The actual implementation of DeltaASP compares exit codes in order to determine whether the observable behavior has changed or not. Instead of passing the name of the answer set solver executable directly to DeltaASP, we pass the name of a wrapper script that calls the answer set solver and returns a specific exit code if the defective behavior occurs, e.g. grep for a specific error message was successful. This approach makes the delta debugger highly flexible. In this way, the concept of observable behavior is not limited to one solver, but can be extended to multiple solvers. For example, considering the classes *invalid models* and *incorrect*, if we observe that two solvers report different results on the same instance, we can use a simple shell script to execute both solvers on the instance passed as argument. If the solvers agree on the result, we return exit code 1 and 0 otherwise. With this technique we delta debugged incorrect results as already proposed in (Brummayer and Biere 2009), but with exactly one trusted solver. Alternatively, one could apply majority voting using multiple solvers.

Notice that, in principle, DeltaASP could reorder rules and rule elements before delta debugging. For example, the rules could be sorted with respect to rule size such that OBO tries to eliminate larger rules first. However, we found out that changing the order of rules and individual rule elements may make the considered failure disappear. Therefore, DeltaASP does not change the original relative order of rules and individual rule elements.

For the experiments, we used a limit of 100 inputs for each class. The results are shown in Tables 4, 5 and 6. Due to page limitations, examples of failure-inducing inputs, and the instances resulting from delta debugging these inputs, can be found in the archive `http://fmv.jku.at/brummayer/fuzz-dd-asp.tar.7z` that contains the full delta debugging results. In Tables 4, 5 and 6, **ins** is the resulting total number of instances delta debugged for each solver, **c** the number of different bug classes, **time** the average delta debugging time in seconds for OBO (obo) and DDMin (ddm), and **size** (resp. **red**) the average size of the resulting instance in bytes (the average reduction in percentages). Notice that, reflecting the reduction in the size of the failure-inducing answer set program, we measure the success of delta debugging in file size. Alternatively, the number of lines or number of rules could be used. However, we found out that these strongly correlate with the file size.

For both of the elimination strategies DDMin and OBO, the average delta debugging times in all categories are less than one minute, and led to an impressively high reduction of at least 98.6%. This clearly shows the effectiveness and overall success of our DeltaASP delta debugging algorithm.

The size reduction achieved by the OBO and DDMin strategies is almost identical for the three program classes NLP, WCP, and DLP. The running times using OBO and DDMin vary more noticeably. Moreover, the more effective strategy depends on the program class. For NLP, DDMin results in making over 50% more calls to the solvers than OBO on average (916 and 572 calls, respectively). This is also reflected in the running times for the strategies on NLP. Thus it seems that a simple one-by-one elimination strategy is the preferred one for NLPs, as the more intricate DDMin makes many ineffective elimination checks. On the other hand, the two elimination strategies give almost identical results on WCPs also time-wise. For DLPs, however, the situation is the opposite to the NLP case: while the difference in the number of solver calls is relatively small, DDMin is over twice as fast as OBO. Comparing this with the NLP case, we believe that the result for DLP is due to the fact that, since disjunctive programs are fundamentally harder to solve than NLPs, it pays off to apply an eager elimination strategy such as DDMin, which results in calling a solver with relatively smaller programs during delta debugging. In particular, if the granularity is rather low, DDMin may eliminate large subsets.

Table 4. Delta debugging results for NLP. Average number of solvers calls is 572 for OBO and 916 for DDMin.

|  |  |  | average | | | | | |
| --- | --- | --- | --- | --- | --- | --- | --- | --- |
| **solver** | **instances** | **classes** | **time** (s) | | **size** (B) | | **reduction** (%) | |
|  |  |  | **obo** | **ddm** | **obo** | **ddm** | **obo** | **ddm** |
| Clasp | 2 | 1 | **32** | 53 | 164 | **98** | 99.0 | **99.4** |
| Cmodels | 4 | 2 | **8** | 12 | **77** | 99 | **98.9** | 98.6 |
| lp2sat | 100 | 1 | 4 | **2** | 90 | **85** | 98.7 | **98.8** |
| lp2diff | 6 | 1 | **6** | 7 | **23** | 24 | 99.2 | 99.2 |
| Smodels-ie | 100 | 1 | 1 | **0** | 4 | 4 | 99.9 | 99.9 |
| Smodels_cc | 5 | 1 | **26** | 88 | 138 | **117** | 99.5 | 99.5 |
| SUP | 100 | 1 | **27** | 37 | 229 | **189** | 99.0 | **99.2** |

Table 5. Delta debugging results for WCP. Average number of solver calls is 311 for OBO and 365 for DDMin.

|  |  |  | average | | | | | |
| --- | --- | --- | --- | --- | --- | --- | --- | --- |
| **solver** | **instances** | **classes** | **time** (s) | | **size** (B) | | **reduction** (%) | |
|  |  |  | **obo** | **ddm** | **obo** | **ddm** | **obo** | **ddm** |
| Clasp | 8 | 2 | 5 | **3** | 40 | 40 | 99.0 | 99.0 |
| Cmodels | 192 | 4 | 6 | **3** | 52 | 52 | 99.2 | 99.2 |
| lp2diff | 8 | 1 | 11 | **5** | 28 | 28 | 99.3 | 99.3 |
| Smodels-ie | 127 | 4 | 5 | **3** | 36 | **22** | 99.1 | **99.5** |
| SUP | 272 | 4 | **11** | 12 | 106 | **83** | 98.9 | **99.2** |

Finally, we note that even non-deterministic solver behavior can be handled by our delta debugging framework. For example, we observed non-deterministic behavior for Smodels-ie which, when run on the same instance multiple times, either crashed with a segmentation fault, terminated with a result, or did not terminate at all. In order to handle such cases in which a solver may not always terminate during delta debugging, time limits can be used. (Here we had to use a time limit (5 seconds) for Smodels-ie.) In more detail, recall that the delta debugger calls the solver (the wrapper script) after each elimination. Each call to a solver is run with a fixed time limit. If the solver does not return a result within the time limit, the shell script returns a specific exit code different from the exit code on the original instance. Then, the delta debugger treats this case as if the elimination has failed, undoes the last elimination, and continues.

Table 6. Delta debugging results for DLP. Average number of solver calls is 508 for OBO and 407 for DDMin.

|  |  |  | average | | | | | |
| --- | --- | --- | --- | --- | --- | --- | --- | --- |
| **solver** | **instances** | **classes** | **time** (s) | | **size** (B) | | **reduction** (%) | |
|  |  |  | **obo** | **ddm** | **obo** | **ddm** | **obo** | **ddm** |
| ClaspD | 139 | 3 | 27 | **9** | 35 | 35 | 99.8 | 99.8 |
| Cmodels | 272 | 7 | 46 | **21** | 75 | **64** | 99.6 | 99.6 |

## 7 Related work

The most closely related work is the fuzz testing and delta debugging approach developed for SMT (satisfiability modulo theories) solvers in (Brummayer and Biere 2009). Our work differs in developing ASP specific fuzzing techniques and, especially, novel delta debugging techniques and strategies in the context of answer set solving. In contrast to our generic black-box approach, solver-specific white-box testing solutions are used in the development process of the DLV solver (Calimeri et al. 2009). In the context of inductive logic programming for data mining, a DDMin-based white-box trace-based delta debugger was developed (Tronçon and Janssens 2006).

As a final note, we want to stress that this work develops debugging techniques for answer set *solvers*, with the aim of developing and providing automated techniques for developing correct solvers. While this work focuses on solver testing and debugging, we note that, when considering applications of ASP, another possible source of errors is the *modeling* phase in which errors may be introduced by either on a conceptual level or through bugs in software which generate answer set programs encoding instances of the application domain. Incorrect modeling can result in answer set programs the answer sets of which do not precisely capture the set of solutions to the original problem instance. Various solutions have been recently proposed for debugging answer set *programs* (Brain and de Vos 2005; Syrjänen 2006; Brain et al. 2007; Gebser et al. 2008) where the aim is to find explanations on why a set of program rules does not describe a correct set of answer sets.

## 8 Conclusions

We developed novel fuzz testing and delta debugging techniques for answer set solver development. The tools provide black-box solutions for more rigorous testing of a wide range of answer set solvers. ASP applications heavily depend on the robustness and correctness of answer set solvers. However, our experimental analysis clearly showed that our fuzz testing tool is able to reveal a variety of different critical defects such as segmentation faults, aborts, infinite loops, incorrect results and invalid answer sets in various state-of-the-art answer set solvers. Moreover, we showed that our delta debugging techniques are very effective in shrinking failure-inducing inputs, which enables efficient debugging of answer set solvers.

As an extension of this work, we are particularly interested in testing and debugging solutions for the non-ground case. As many of the current state-of-the-art solvers heavily depend on the robustness and correctness of grounders, we find this an interesting and important aspect of future work.